\documentclass{article}
\usepackage{ijcai23}

\usepackage{times}
\usepackage{soul}
\usepackage{url}
\usepackage[hidelinks]{hyperref}
\usepackage[utf8]{inputenc}
\usepackage[small]{caption}
\usepackage{graphicx}
\usepackage{amsmath}
\usepackage{amsthm}
\usepackage{booktabs}
\usepackage{algorithm}
\usepackage{algorithmic}
\usepackage[switch]{lineno}
\usepackage{stfloats}
\usepackage{float}
\usepackage[section]{placeins}
\usepackage{xcolor}

\title{Addressing Variable Dependency in GNN-based SAT Solving}

\author{
Zhiyuan Yan$^1$
\and
Min Li$^2$\and
Zhengyuan Shi$^2$\and
Wenjie Zhang$^3$ \and
Yingcong Chen$^1$ \And
Hongce Zhang$^1$
\affiliations
$^1$Hong Kong University of Science and Technology(Guangzhou)\\
$^2$The Chinese University of Hong Kong \\
$^3$Peking University
\emails
zyan760@connect.hkust-gz.edu.cn$^1$,
\{mli,zyshi21\}@cse.cuhk.edu.hk$^2$,
zhang\_wen\_jie@pku.edu.cn$^3$,
yingcong.ian.chen@gmail.com$^1$,
hongcezh@ust.hk$^1$
}

\begin{document}

\maketitle

\begin{abstract}
Boolean satisfiability problem (SAT) is fundamental to many applications.
Existing works have used graph neural networks (GNNs) for (approximate) SAT solving.
Typical GNN-based end-to-end SAT solvers predict SAT solutions \emph{concurrently}. 
We show that for a group of symmetric SAT problems, the concurrent prediction is guaranteed to produce a wrong answer because it neglects the dependency among Boolean variables in SAT problems. 
We propose AsymSAT, a GNN-based architecture which integrates recurrent neural networks 
to generate dependent predictions for variable assignments. The experiment results show that dependent variable prediction extends the solving capability of the GNN-based method as it improves the number of solved SAT instances on large test sets.
\end{abstract}

\section{Introduction}
Boolean satisfiability problem (or the SAT problem) is to decide if there exists a set of 0-1 (false/true) assignments to the Boolean variables in a given Boolean formula such that the formula evaluates to 1 (true). Solving the SAT problem is at the core of many applications, for example, the electronic design automation (EDA), where the growing complexity of hardware 
demands for more intelligent design and verification tools \cite{brand1993verification}.
One of the classic applications of SAT solving in EDA is combinational logic equivalence checking, where two circuits for comparison are composed into one miter circuit~\cite{gupta2006sat}.
This miter circuit can be further converted into a conjunctive norm form (CNF) formula. If the formula is satisfiable, then the two circuits are not equivalent. 
In this paper, we refer to solving the Boolean formula in CNF as the SAT problem.
If the problem input is directly in the circuit form, then we call it the Circuit-SAT problem.\looseness=-1  


Traditionally, SAT and Circuit-SAT problems are solved by heuristic searching algorithms, for example, the Conflict-Driven Clause Learning (CDCL) algorithm~\cite{marques1999grasp}, implemented in solvers like MiniSAT~\cite{sorensson2005minisat} and Glucose~\cite{audemard2014glucose}.
While SAT problems are NP-complete, heuristics play an important role in speeding up the algorithms. Many effective heuristics have been proposed so far.
For example, the Variable State-Independent Decaying Sum (VSIDS) method was brought up to aid variable selection in the search process~\cite{moskewicz2001chaff}. 
 Those heuristic algorithms have led to a huge improvement in computational efficiency. However, they are also limited by the greedy nature of their strategy and therefore, could be sub-optimal in certain cases~\cite{shi2021transformer}. 

Recently, there has been a growing interest in applying machine learning methods to solve combinatorial problems, including the SAT problem~\cite{duan2022augment}.
With the help of deep neural networks (DNN), end-to-end models can learn the expected outcome from the problem description. For example,  \cite{selsam2018learning} proposed the NeuroSAT architecture to approach SAT solving as a binary classification problem.
NeuroSAT treats the input CNF as a bipartite graph and uses the graph neural network (GNN) to predict whether the set of CNF clauses is satisfiable or not. Compared to the existing heuristic searching method, NeuroSAT requires no pre-defined heuristics.
\cite{amizadeh2018learning} considered the scenario when the problem formulation is in the form of a circuit (namely the Circuit-SAT problem). As the circuit is by nature a graph, the GNN-based architecture also applies here. Unlike NeuroSAT which requires extra steps to decode a satisfying assignment from a SAT prediction, the neural Circuit-SAT method in~\cite{amizadeh2018learning} directly predicts the satisfying assignment. These existing GNN-based end-to-end methods have demonstrated the potential of using machine-learning techniques in SAT solving.

Although prior works showed that it is possible to predict SAT solutions from the graph structure of the problem formulation, we identify an important flaw in these existing GNN-based end-to-end SAT solving methods: there exists a set of satisfiable CNF formulas (correspondingly, a set of satisfiable circuits for the Circuit-SAT problem) whose satisfying assignments cannot be learned by the existing methods in \cite{selsam2018learning} or \cite{amizadeh2018learning}. 
These CNF formulas (or circuits) have symmetric formulations, but their solutions are asymmetric. Here, symmetry means swapping a pair of variables resulting in exactly the same CNF formula (or circuit). However, a satisfying assignment requires that the two symmetric variables must take different values.
A simple example is the formula (or the circuit) that encodes ``$a \oplus b$'', where $a$ and $b$ are two Boolean variables and ``$\oplus$'' is the XOR operation. Due to the commutativity of XOR, ``$b\oplus a$'' is equivalent to ``$a \oplus b$''. However, a satisfying assignment requires $a$ and $b$ to take different values because only ``$0 \oplus 1$'' or ``$1 \oplus 0$'' results in $1$.  
Existing methods predict Boolean assignments for $a$ and $b$ concurrently. The concurrent prediction on each variable solely depends on the graph embedding of the variable node. Meanwhile, $a$ and $b$ share the same embedding because they are symmetric in CNF or circuit form. Therefore, the predictions for  $a$ and $b$ inevitably become the same, which is clearly not a satisfying variable assignment for the formula.

We attribute this problem to the concurrent prediction in existing model that does not factor in variable dependency in SAT problems.
Variable dependency refers to the fact that the prediction of one variable assignment will affect other variables. In the ``$a \oplus b$'' example, if $a$ is assigned to be 0, then $b$ must be 1, despite that $a$ and $b$ are symmetric. 
To address this problem, we introduce a recurrent neural network (RNN) in the SAT assignment decoding layer.
With the help of RNN, later predictions will be able to ``remember'' prior variable assignments.  We call our new model AsymSAT, because it can produce asymmetric SAT solutions given symmetric SAT problems.
We show by experiments that this small change can significantly improve GNN-based SAT solving.

Overall, our main contributions in this paper are:
\begin{itemize}
     \item We identify the need of addressing variable dependency in the existing GNN-based end-to-end SAT solving methods. 
     \item We propose an improvement to the neural network architecture to 
     take dependency among variables into consideration. Our AsymSAT model uses  RNN to make sequential predictions of SAT solutions.
     \item We demonstrate that with this small change, AsymSAT achieves a higher accuracy in SAT and Circuit-SAT solving compared to prior works. In addition to the end-to-end machine-learning-based SAT solving,  the idea of sequential predictions proposed in this paper could potentially also benefit hybrid SAT solvers that take machine-learning as one searching heuristic.\looseness=-1
\end{itemize}

The paper is organized as follows: the next section provides a background on SAT problems and the existing GNN-based end-to-end SAT solving methods. Section~\ref{symmetric} highlights the variable dependency in SAT solving. 
Section~\ref{our method} introduces our improvement to the GNN architecture for SAT solving, followed by experiments in Section~\ref{experiment} and the related works in Section~\ref{related work}. 
Finally,  Section~\ref{conclusion} concludes the paper.

\section{Background}
\label{background}
%
\subsection{Boolean Satisfiability Problem}

The Boolean satisfiability problem talks about propositional logic with Boolean variables and Boolean operators like ``and''($\land$), ``or''($\lor$), ``not'' ($\lnot$). 
The problem is to decide whether there exist assignments to the Boolean variables so that a given propositional logic formula evaluates to true under such assignment. 
The expected answer is either satisfiable (SAT) or unsatisfiable (UNSAT). In case the formula is satisfiable, we also expect to know the satisfying variable assignment. 

In Boolean satisfiability problem, the input is usually a Boolean formula in the conjunctive norm form (CNF). In CNF, the variables and their negations are called the literals, which are first connected with disjunctions ($\lor$) to form the clauses. The clauses are then connected with conjunctions ($\land$). 
An arbitrary Boolean formula can always be converted into CNF using Boolean algebra, though the number of clauses may exponentially blow up. Alternatively,  Tseitin transformation~\cite{tseitin1983complexity} converts a formula into an equi-satisfiable CNF, whose size is linear to the number of operators in the original one. Therefore, it is the preferred solution of CNF conversion in modern SAT solving.

Besides CNF, the input of SAT could also be a circuit, which is essentially a directed acyclic graph (DAG) with each node representing either a circuit input, a logic gate or the circuit output. The goal is to find an input combination that causes the circuit output to become 1. The circuit  form is more natural in some applications (for example, EDA). 
Note that the circuit form  can be converted into a CNF formula by introducing new variables, 
and the number of clauses and variables in the converted CNF is proportional to the number of graph nodes in the circuit DAG~\cite{prestwich2009cnf}.
On the other hand, a CNF formula can also be converted back into a circuit form, as presented in \cite[Appendix C]{amizadeh2018learning}.

\subsection{Solving SAT Problems by Graph Neural Networks}\label{NeuroSAT}
Selsam \textit{et al.} proposed using a graph neural network to decide whether a Boolean formula in CNF  is satisfiable or not~\cite{selsam2018learning}.
This is an end-to-end method, where the problem input is a bipartite graph with nodes representing either a literal or a clause in CNF. The only output of the neural network is a single bit representing the satisfiability of the formula. An edge between a literal node and a clause node in the bipartite graph indicates that the literal is contained in the clause. Each node has an initial embedding, which is updated by iterations of message-passing between literals and clauses. Finally, the literal embeddings are sent to a multi-layer perceptron (MLP) to generate a vote. The averaged vote from all literals  is used to decide satisfiability.
The paper also demonstrated it is possible to decode satisfying assignments from the literal embeddings via k-means clustering.


\subsection{Solving Circuit-SAT Problems by Graph Neural Networks}
When the problem is formulated in the circuit form, we automatically have a graph structure. Graph neural network can  be similarly applied in Circuit-SAT problems. 
Note that any logic function can be implemented with only AND gates and NOT gates. A circuit graph made up of only these two types of gates forms an And-Inverter-Graph (AIG), which is commonly used in the EDA community. 
Amizadeh \textit{et al.} developed a GNN-based model, named DG-DAGRNN  to solve Circuit-SAT problems by machine learning~\cite{amizadeh2018learning}.
The types of graph nodes (whether a node is a logic gate or a circuit input) are encoded as one-hot vectors, which are the inputs to the GNN model. Similar to NeuroSAT, message-passing generates node embeddings which are used to predict variable assignments. 
A message-passing iteration consists of one forward pass from circuit input nodes to the only circuit output node and one backward pass in the reversed order.
Compared to NeuroSAT,
message-passing in DG-DAGRNN follows the topological order and sequentially update node embeddings, whereas NeuroSAT updates all node embeddings concurrently.

Both NeuroSAT and DG-DAGRNN predict SAT solutions concurrently without considering dependency among variables. This results in a fundamental weakness: they are unable to predict the correct solutions for certain symmetric graphs, as explained in the next section.

\section{Variable Dependency in SAT Solving}
\label{symmetric}
Generally speaking, SAT and Circuit-SAT solving must consider variable dependency.
In other words, they must ``remember'' what predictions have been made so far. A simple example is the 2-input XOR ($x \oplus y$). Here, $x$ and $y$ are symmetric --- if we swap them, we will get exactly the same formula because XOR is commutative. However, we must assign different values to $x$ and $y$ in order to get a 1 as the result. 
If $x$ has been assigned as 1, then $y$ must be 0. This is the dependency between these two variables.

\begin{figure*}[h]
\centering
\includegraphics[width=0.75\textwidth]{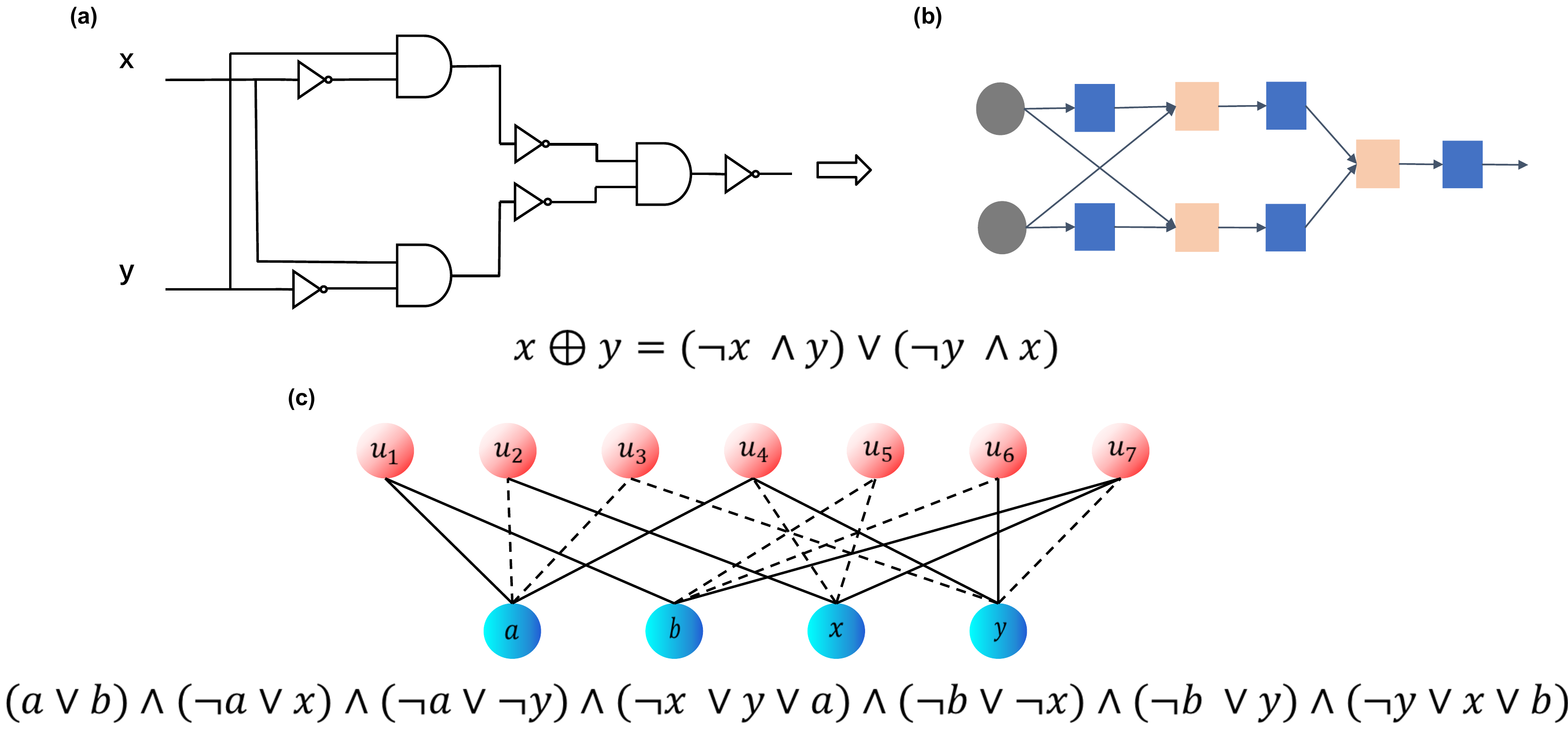}
\caption{(a) XOR implemented by AIG; (b) the DAG representation of (a); (c) the equi-satisfiable CNF with additional variable $a$ and $b$, and the corresponding bipartite graph of XOR (here  dotted line means the variable is negated in the clause).}
\label{fig:XOR}
\end{figure*}

Symmetry natually exists in many SAT problems. Sometimes, it is part of the formula that is symmetric --- for example, $( x \oplus y ) \land z$.
When converted to AIG or CNF, 
a symmetric formula like $x \oplus y$ will result in a symmetric AIG or CNF, as shown by Figure~\ref{fig:XOR}.
It is not hard to see, symmetric nodes have symmetric predecessors and successors. Therefore, when GNN-based SAT solvers use message-passing to encode the graph structure,  symmetric nodes
will have the same node embeddings,  unless they are distinguished by initialization. However, 
pure random initialization for all nodes provides no extra information for the neural network to distinguish the symmetric ones. 
On the other hand, a bias in initialization would introduce artefact that does not generalize. Therefore, prior works~\cite{selsam2018learning,amizadeh2018learning,zhang2020nlocalsat} all used equal initial embeddings and therefore, they would not be able to distinguish symmetric nodes when predicting SAT assignments. We accompany our argument on random initialization with experimental results  in the appendix.
%

When individual node embeddings are directly used to predict variable assignments without considering the dependency among them, the inferred assignments will always be the same for the pair of symmetric nodes. As we have shown by the 2-input XOR example, some symmetric formulas reject equal variable assignments as their satisfying solutions. Therefore, NeuroSAT and DG-DAGRNN in \cite{selsam2018learning} and \cite{amizadeh2018learning} are  unable to deal with these symmetric SAT or Circuit-SAT problems.
We argue that a GNN-based SAT solver should \emph{sequentially} predict variable assignments in order to take variable dependency into consideration. This is achieved by a recurrent neural network added in our model, explained in the next section.

\section{Our Methods} 


 In this section, we explain our approach where RNN is used for dependent predictions.
Specifically, we focus on the Circuit-SAT problem because a CNF formula for SAT problems can be converted into a circuit structure. 
 We formulate solving Circuit-SAT problems as a supervised learning process as the  following.

\label{our method}
 \subsection{Problem formulation}
 \label{sec:our-method}
\textbf{Problem input.} 
We expect the problem input to be a DAG representing the structure of the circuit. As discussed in Section~\ref{background}, we only need to consider circuits made from AND gates and NOT gates. Each node in the DAG has an one-hot input feature vector that indicates the type of the node.
There are in total three types: the primary inputs, AND gates and NOT gates.
 Formally, we expect the problem input to be the form of $G =<V_{G}, N_{G}, E_{G}>$, where $ V_{G}$ is a set of circuit nodes, $ N_{G}$ is a function that maps each node to its type, and $ E_{G}$  is the set of directed edges of the circuit graph. An edge between two nodes means that there is a wire connection from a logic gate or a circuit input to another gate.\looseness=-1


\textbf{Problem output.} The machine learning model should predict a 0-1 assignment for each circuit input node. We denote the assignments as  $L\in \left\{0,1\right\}^i$ and $i$ is the number of circuit input nodes.
Each instance in the dataset is in the form of $(G, L)$, where the 0-1 assignments are generated by an external SAT solver that works as the oracle.

 \subsection{The Proposed GNN Architecture}
 \label{GNN Architecture}
 
 In the high-level, we would like to build a machine-learning model that learns the mapping from a circuit graph to the 0-1 assignment on input nodes: 
 $f : G \to L$. There are plenty of existing GNN models that are designed to handle input data organized as a graph~\cite{yolcu2019learning,selsam2018learning,selsam2019neurocore}. 
 There, each graph node is associated with a vector (the hidden state vector) which eventually represents some structural information around the node. Nodes exchange their knowledge of the graph structure by sending messages to their neighbors, and the hidden states will be gradually updated. The propagation of information is referred to as the message-passing mechanism, which essentially embeds the information about the graph structure into the hidden states.

\textbf{Graph embedding layers.}
When it comes to the implementation of message passing, there are various choices, for example,  which direction the message flows towards, how to aggregate messages from several nodes, what is the order of hidden state updates. Therefore, different variants of message-passing can be implemented. In this work, we build upon the DAG-RNN framework~\cite{shuai2016dag} to create a GNN architecture for sequential variable assignment prediction.

To better explain our GNN architecture, we introduce the following notations. Each graph node $v\in V_{G}$ is associated with a $d$-dimensional hidden state vector ${x}_v$, which is iteratively updated based on the messages from neighboring nodes.
During message-passing,
we distinguish the nodes that reach $v$ following a directed edge (in other words, the predecessors) from those that leaves $v$ (the successors). We only use the messages from predecessors in the forward pass, and likewise, the successors in the backward pass. The incoming messages are aggregated by an aggregator function $\mathcal{A}$, which is invariant to permutation of the elements in the input set. Finally, the aggregated message is used to update the hidden state of $v$ by a 
standard GRU function $GRU \left(\cdot \right)$~\cite{cho2014learning}.

In AsymSAT, message passing follows the topological order. In the forward pass, messages flow from circuit input nodes (which have no predecessors) to the only circuit output node (which has no successors). The hidden state vectors are updated sequentially. In the backward pass, messages flow from the circuit output node to the circuit input nodes. 
In each pass, the hidden state vectors are updated according to the following rule:
\begin{equation}\label{eq:node-update}
{x}_v^{(k+1)} := GRU \left( {p}_v, 
    \mathcal{A}
    \left(\left\{
         {m}_n^{(k)} | n\in \mathcal{N} \left( v\right)
    \right\}\right)
\right)
\end{equation}
Initially, ${p}_v=N_{G} \left(v \right)$,
 which is the node type vector of node $v$. 
So in the first forward pass, the type of a node is encoded into the hidden state vector. In all remaining passes, ${p}_v={x}_v^{(k)}$, which is the  hidden state vector resulted from the previous pass. We use three separate GRUs: $GRU_{init} \left(\cdot \right)$, $GRU_{f} \left(\cdot \right)$, $GRU_{b} \left(\cdot \right)$. Among the three, $GRU_{init} \left(\cdot \right)$ is only used in the first forward pass.
$GRU_{f} \left(\cdot \right)$ is used for all remaining forward passes. $GRU_{b} \left(\cdot \right)$ is used in the backward passes. We call one forward pass followed by one backward pass as an iteration.
In Equation~\ref{eq:node-update}, $\mathcal{N} \left( v\right)$ is either the predecessors or the successors of $v$.  Their hidden state vectors are encoded into  messages ${m}_n^{(k)}$ by a learnable function $\mathcal{M} : {x}_n^{(k)} \to {m}_n^{(k)}$.

Our graph embedding layers share some similarities with \cite{amizadeh2018learning} as both are built upon DAG-RNN. The difference here is mainly in the computation between two iterations. We use two GRUs for forward passes, because the size of a hidden state vector is different from that of the node type vector, 
whereas \cite{amizadeh2018learning} introduced a function to project hidden state vectors into the space of node type vectors after each iteration to keep the same dimensionality and use the same GRU. 
We argue that the projection could potentially introduce a loss of information and therefore, we employ two separate GRUs in the forward pass: $GRU_{init} \left(\cdot \right)$, $GRU_{f} \left(\cdot \right)$ to handle either the node type vector or a hidden state vector from the previous pass.








\textbf{SAT assignment decoding layers.} 
As discussed earlier, we would like to predict the Boolean assignments on the circuit input nodes \emph{sequentially}. In this node-level prediction, we map
a sequence of hidden state vectors of the circuit input nodes $X = \left( {x}_{i_1} , {x}_{i_2} , {x}_{i_3} , ...\right)$ to a sequence of input assignment $L$. After iterations of message passing, these hidden state vectors encode the information related to the structure of the graph. If two input nodes are symmetric with respect to each other, we expect that their hidden state vectors will be the same. 
If we individually use each of these vectors to decode a 0-1 assignment (namely, the concurrent prediction), the symmetric nodes will certainly map to the same variable assignment. As we have discussed in Section~\ref{symmetric}, 
SAT solutions must take variable dependency into consideration, 
therefore, in our model, we need to associate variable assignments of the same SAT problem.

In our AsymSAT, we use a recurrent neural network (referred to as the $\mathcal{R}$ layer)  to generate sequential predictions on variable assignments, so that the model output on a certain circuit input node depends on the predictions of other nodes. 
We make this $\mathcal{R}$ layer bi-directional to account for dependencies from both sides. A subsequent MLP will work as a selector to decide which direction is more preferred.
Sequential prediction mimics classic (non-machine-learning-based) SAT solvers. These classic SAT solvers like GRASP~\cite{marques1999grasp} or MiniSAT~\cite{sorensson2005minisat} pick decision variables one after another. 
Regarding the aforementioned XOR example, we expect this RNN layer will be able to learn to predict different variable assignments for the two symmetric variables after training with such examples.\looseness=-1

\begin{figure*}
\centering
\includegraphics[width=0.87\textwidth]{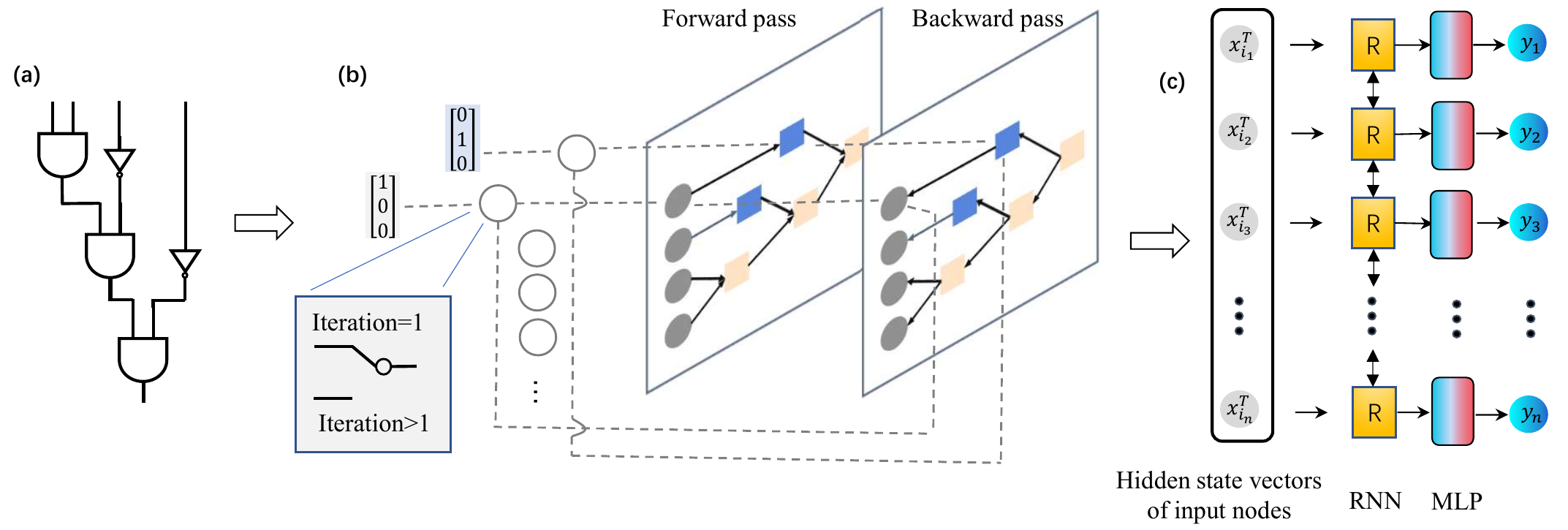}
\caption{(a) An example of an AIG circuit, (b) the graph embedding layers (for simplicity, we only draw the connections for two nodes.) (c) the SAT assignment decoding layers}
\label{fig:arch}
\end{figure*}

As a summary, we show the overall architecture of our AsymSAT model in Figure~\ref{fig:arch}.




\subsection{Training}
\label{training}

For AsymSAT, we apply the supervised learning method.
We consider SAT solution prediction as a labeling problem --- giving 0-1 labels to each of the circuit input nodes.
We use Cross-Entropy for the loss function, denoted as:  
\begin{equation}
\begin{split}
Loss  =&   - \Sigma_{i=1}^n [
g\left.{(}v_{i}\right.)log\left.(P\left.{(}v_i=1\right.{)}\right.{)}
+ \\
&\left.{(}1-g\left.{(}v_{i}\right.{)}\right.{)}log\left.{(}P\left.{(}v_i=0\right.{)}\right.{)}]
\end{split}
\end{equation}
where $g(v_{i})$ is the ground truth of the SAT assignment on $v_{i}$ generated by an oracle SAT solver, $P(v_i=1)$ and $P(v_i=0)$ are the predicted probability of $v_i$ to be 1 or 0.

\begin{table*}[!htb]
\caption{Percentage of the symmetric circuit problem solved}
\begin{center}
\setlength{\tabcolsep}{1.1mm}{
\begin{tabular}{rrrrr}
\multicolumn{1}{c}{\bf AsymSAT w. LSTM}   &\multicolumn{1}{c}{\bf AsymSAT w. GRU} &
\multicolumn{1}{c}{\bf AsymSAT w.o. $\mathcal{R}$ layer} &
\multicolumn{1}{c}{\bf NeuroSAT} &\multicolumn{1}{c}{\bf DG-DAGRNN}
\\ \hline \\
$100.00\%$   & $100.00\%$   &
$0.00\%$  &$0.00\%$
&$0.00\%$

\label{Symmetric circuit result}
\end{tabular}
}
\end{center}
\end{table*}

\section{Experimental Evaluation}
\label{experiment}
\subsection{Data preparation} 
\begin{table*}[htbp]
\caption{Solution rate for the $SR(n)$ problems}
\label{SR result}
\begin{center}
\setlength{\tabcolsep}{0.9mm}{
\begin{tabular}{lllllllllll}
\multicolumn{1}{c}{}&\multicolumn{1}{c}{\bf $SR(3)$}  &\multicolumn{1}{c}{\bf $SR(4)$} &\multicolumn{1}{c}{\bf $SR(5)$}
&\multicolumn{1}{c}{\bf $SR(6)$}
&\multicolumn{1}{c}{\bf $SR(7)$}
&\multicolumn{1}{c}{\bf $SR(8)$}
&\multicolumn{1}{c}{\bf $SR(9)$}
&\multicolumn{1}{c}{\bf $SR(10)$} 
\\ \hline \\
{\bf AsymSAT} &$98.30\%$         &$100.00\%$
&$93.23\%$   &$94.51\%$ &$81.56\%$ &$82.90\%$&$88.95\%$&$85.45\%$


\\

{\bf NeuroSAT} &$87.70\%$         &$74.47\%$
&$63.10\%$   &$59.57\%$ &$52.94\%$ &$48.40\%$&$49.73\%$&$43.82\%$

\\

{\bf DG-DAGRNN} &$10.21\%$         &$15.23\%$
&$5.21\%$   &$1.83\%$ &$8.38\%$ &$5.70\%$&$4.07\%$&$4.24\%$
\end{tabular}
}
\end{center}
\end{table*}
\begin{table*}[t]
\caption{Solution rate for the $V(n)$ problems}
\label{aigen result}
\begin{center}
\setlength{\tabcolsep}{0.9mm}{
\begin{tabular}{lllllllllll}
\multicolumn{1}{c}{}&\multicolumn{1}{c}{\bf $V(3)$}  &\multicolumn{1}{c}{\bf $V(4)$} &\multicolumn{1}{c}{\bf $V(5)$}
&\multicolumn{1}{c}{\bf $V(6)$}
&\multicolumn{1}{c}{\bf $V(7)$}
&\multicolumn{1}{c}{\bf $V(8)$}
&\multicolumn{1}{c}{\bf $V(9)$}
&\multicolumn{1}{c}{\bf $V(10)$} 
\\ \hline \\
{\bf AsymSAT} &$81.58\%$         &$67.50\%$
&$72.50\%$   &$55.50\%$ &$52.50\%$ &$60.00\%$&$45.00\%$&$47.50\%$

\\


{\bf NeuroSAT} &$0.025\%$         &$0.00\%$
&$0.00\%$   &$0.00\%$ &$0.00\%$ &$0.00\%$&$0.00\%$&$0.00\%$
\\

{\bf DG-DAGRNN} &$35.00\%$         &$47.50\%$
&$47.50\%$   &$45.00\%$ &$30.00\%$ &$37.50\%$&$37.50\%$&$32.50\%$
\end{tabular}
}
\end{center}

\end{table*}
\begin{table*}[h]
\caption{Solution rate for the larger $V(n)$ problems}
\label{generalized result aigen}
\begin{center}
\setlength{\tabcolsep}{0.6mm}
{
\begin{tabular}{lrrrrr}
\multicolumn{1}{c}{}&\multicolumn{1}{c}{\bf $V(11)$} &\multicolumn{1}{c}{\bf $V(12)$}&\multicolumn{1}{c}{\bf $V(13)$}&\multicolumn{1}{c}{\bf $V(14)$}&\multicolumn{1}{c}{\bf $V(15)$}
\\ \hline \\
{\bf AsymSAT trained on SR(3..10)} &$45.00\%$ &$60.00\%$ &$45.00\%$ &$45.00\%$ &$52.50\%$ \\
{\bf AsymSAT trained on SR(3..10) + V(3..8)} &$47.50\%$ &$47.50\%$ &$45.00\%$ &$60.00\%$ &$57.50\%$ \\




\end{tabular}
}
\end{center}
\end{table*}
We prepare three datasets in total: the small-scale symmetric circuit examples, medium-size CNF formulas, and large random circuits with more than 1K logic gates.

\textbf{Small-scale symmetric AIG with asymmetric solutions.}
We manually construct 10 circuits with no more than 3 inputs. Within each circuit, there are at least two input nodes that are symmetric but require distinct assignments. 
We intentionally keep this training set small. If NeuroSAT and DG-DAGRNN are capable of handling symmetric circuits with asymmetric SAT solutions, they should easily reach a high training accuracy on this small dataset. However, our experiment result later will show that they are unable to predict any SAT solutions for this dataset. 

\textbf{Medium-size randomly generated CNF formulas.} 
We generate random CNF formulas in the same way as described by \cite{selsam2018learning}. We refer to this dataset as the $SR(n)$ problem, where $n$ is the number of variables. CNF formula for $SR(n)$ problems can be converted into the circuit form 
using the principle of Shannon's Decomposition as suggested by  \cite{amizadeh2018learning}.

\textbf{Large randomly generated AIGs.} 
We generate random AIGs using the AIGEN tool~\cite{jacobs2021aigen}, which was designed to create random test circuits to check and profile the EDA tools. 
By default, AIGEN generates sequential logic circuits (those with storage elements). We extract the combinational logic circuits from the sequential logic circuits. We refer to this dataset as the $V(n)$ problem, where $n$ stands for the number of circuit input nodes.  $V(n)$ problems can be converted into CNF using Tseitin transformation.
Compared to $SR(n)$ problems, $V(n)$ is a nontrivial dataset even when $n$ is relatively small. For example, each $V(10)$ problem has more than 1K logic gates on average.
The corresponding CNF formulas contain more than 1K variables, which is much larger than the largest dataset $SR(40)$ used in the prior work~\cite{selsam2018learning}.

\subsection{Experimental setup and result}

The dimension on the outcome of the $\mathcal{R}$ layer is 10 and we use the Adam optimizer during training process. 
For NeuroSAT, and DG-DAGRNN, we follow the same configurations as described in \cite{selsam2018learning} and \cite{amizadeh2018learning}. 
To our best knowledge, the source code for the original DG-DAGRNN model is not publicly available. We build this model following the instructions in  \cite{amizadeh2018learning}. 
We train and test all three models on a server with two NVIDIA GeForce RTX 3090 GPUs. 





\subsubsection{Experiments for AsymSAT Configurations}
\label{discussion}

\textbf{Effectiveness of the RNN decoding layer.}  In this experiment, we train our AsymSAT model, the NeuroSAT model and the DG-DAGRNN model on the same 10 symmetric circuits and measure the training accuracy. We use two configurations for our AsymSAT model: one uses LSTM and the other uses GRU in the bi-directional RNN layer (the $\mathcal{R}$ layer).
 We also add one case of removing the  $\mathcal{R}$ layer in AsymSAT as  comparison.
  In this experiment, we set the learning rate of AsymSAT models as $10^{-3}$ and the number of iterations as 5. 
  Table~\ref{Symmetric circuit result} illustrates the result for the symmetric circuits on five different models. Just as we discussed in Section~\ref{symmetric},  DG-DAGRNN and NeuroSAT cannot break the tie in symmetric circuits or symmetric CNF formulas. 
 And there is no way to train these two models on this dataset. Thanks to the $\mathcal{R}$ layer  we introduced, our AsymSAT model can reach a solution rate of $100.00\%$ with either LSTM or GRU in the $\mathcal{R}$ layer. This shows the effectiveness of the $\mathcal{R}$ layer for symmetric circuits.






 \textbf{Setting the number of iterations.} In this experiment, we study the effect of changing the number of iterations on our network. 
We set the iteration to be 5, 10, 15, and 20, respectively, and test on a mixed dataset with instances from $SR(3)$ to $SR(10)$. We can see for AsymSAT with GRU, increasing the number of iteration from 5 to 10 will greatly improve the accuracy, then the solving rate barely increases for more iterations. 
AsymSAT with LSTM shows a similar result, while it peaks at around 15 iterations.
%
%
It seems that AsymSAT with LSTM may have a higher potential to achieve a better accuracy. Therefore, in the following experiments on SAT solving, we mainly use AsymSAT with LSTM for comparison. 
\begin{table}[h]
\caption{Solution rate under different iterations}
\label{iteration result}
\begin{center}
\setlength{\tabcolsep}{0.9mm}{
\begin{tabular}{lllll}
\multicolumn{1}{c}{\# of iterations}&\multicolumn{1}{c}{5}  &\multicolumn{1}{c}{ 10} &\multicolumn{1}{c}{ 15}&\multicolumn{1}{c}{ 20}
\\ \hline \\
{\bf AsymSAT w. GRU } &$80.63\%$         &$90.32\%$
&$90.25\%$ &$89.11\%$ \\ 
{\bf AsymSAT w. LSTM } &$79.76\%$         &$90.45\%$
&$93.07\%$ &$91.80\%$ 
\end{tabular}
}
\end{center}
\end{table}

\subsubsection{Experiments for SAT solving}

In the following experiments, we compare the three models: our AsymSAT model with bi-directional LSTM in the $\mathcal{R}$ layer, the NeuroSAT model, and the DG-DAGRNN model. 
We measure the performance using solution rate rather than the accuracy of  predicting satisfiability.
Solution rate is defined as the percentage of problems on which the network is able to predict one satisfying assignment.

We admit that for different models, sometimes it is hard to make an absolutely fair comparison.
As AsymSAT is supervised under the multi-bit SAT solution, whereas NeuroSAT is one-bit SAT/UNSAT supervision, AsymSAT 
is much more training-data-efficient and it can converge several orders of magnitute faster than NeuroSAT, although their parameter counts are roughly in the same scale (around 200K).
Actually, the authors of NeuroSAT reported that it would take $10^7$ SAT problems to train a full-fledged NeuroSAT~\cite{neurosat-issue}. 
Although our training set is much smaller due to limitations of computing resources, we didn't observe the overfitting problem in the experiments judging from the training loss.
\looseness=-1

\textbf{Comparison on the $\boldsymbol S\boldsymbol R \boldsymbol( \boldsymbol n\boldsymbol)$ problems.}
We use $8K$ $SR(n)$ problems sampled uniformly from $SR(U(3,8))$ to train the three models. The test set contains $1.5K$ $SR(n)$ problems ($n$ is from 3 to 10). For AsymSAT and DG-DAGRNN, CNF formulas are first converted into circuits to serve as the model input. 
Table~\ref{SR result} summarizes the performance measured on the $SR(n)$ problem. The result shows that AsymSAT model has a better performance compared to NeuroSAT and DG-DAGRNN on this dataset. Overall, AsymSAT can reach more than $90\%$ solution rate (averaged across $SR(3)$ to $SR(10)$), while NeuroSAT can only reach $60\%$. 
Furthermore, we supply more experimental results regarding larger $SR(n)$ problems. When trained from $SR(3)$ to $SR(10)$, AsymSAT outperforms NeuroSAT on $SR(20)$ , $SR(40)$, $SR(60)$ and $SR(80)$. The result is shown in Table~\ref{generalized result}. 
In our experiment, the performance of DG-DAGRNN is non-competitive to the other two.
We conjecture that the unsupervised learning method in DG-DAGRNN suffers from the vanishing gradient problem if trained on circuits converted from CNF.
We provide a detailed analysis of DG-DAGRNN in the appendix. 

\textbf{Comparison on the $\boldsymbol V \boldsymbol( \boldsymbol n\boldsymbol)$ problems.} The training data is a mixture of $8K$ $SR(n)$ problems, ($n$ ranges from 3 to 10), and $1.2K$ $V(n)$ problems ($n$ ranges from 3 to 8).  The test set is 320 $V(n)$ problems ($n$ ranges from 3 to 10). 
Note that $V(n)$ is a nontrivial dataset. On average, each $V(10)$ problem  has around 1K AND gates, more than those in circuits converted from the $SR(10)$ problems (which each contain just about 200 AND gates).
Even for the $SR(40)$ problems, there are only approximately 600 - 800 AND gates per input. Therefore, $V(n)$ problems can also demonstrate the generalization capability of the tested models. Although the indexing number $n$ is relatively smaller in the $V(n)$ dataset, there are plenty of logic gates in each circuit. 
These logic gates will add up to the number of variables and clauses after Tseitin transformation.
 Therefore, the converted CNF inputs are challenging for NeuroSAT. This explains the poor performance of NeuroSAT in Table~\ref{aigen result}. We also show the generalizability of AsymSAT on the $V(n)$ dataset. On larger $V(n)$ problems, for example, $V(15)$, which is about 128x the size of $V(8)$, AsymSAT still maintains a solution rate around 50\%. It is not significantly affected by reducing the training set to only the $SR(n)$ problems as shown by Table~\ref{generalized result aigen}.

In summary, our AsymSAT model is capable of breaking the tie in symmetric circuits and it achieves a higher solution rate in comparison with NeuroSAT and DG-DAGRNN on both medium-size CNF problems and large-size Circuit-SAT problems. This shows the effectiveness of using RNN to account for variable dependency in GNN-based SAT solving.

\section{Related Works}
\label{related work}





\subsection{SAT Solvers}
There are two main categories of machine-learning-based SAT solvers: the end-to-end SAT solvers and the solvers using machine-learning as just the heuristics. NeuroSAT~\cite{selsam2018learning}, DG-DAGRNN~\cite{amizadeh2018learning} and our AsymSAT all belong to the first category, where machine learning methods are used to directly predict the SAT outcome. In the second category, machine learning methods only serve as a heuristic, guiding the classic algorithms. For example, NeuroCore~\cite{selsam2019neurocore} used GNN to compute scores for variable selection in SAT solving and NLocal-SAT~\cite{zhang2020nlocalsat} used GNN to predict one potential solution as the starting point of the stochastic local search (SLS) process. There are also other techniques to support SAT solving. For example, QuerySAT proposed to use multiple SAT queries to increase accuracy~\cite{ozolins2021goal}, and \cite{li2022deepsat}  suggested it is helpful to transfer SAT problems from different application domains to a unified underlying distribution.  
\begin{table}[h]
\caption{Solution rate for the larger $SR(n)$ problems}
\label{generalized result}
\begin{center}
{
\begin{tabular}{lrrrr}
\multicolumn{1}{c}{}&\multicolumn{1}{c}{\bf $SR(20)$} &\multicolumn{1}{c}{\bf $SR(40)$}&\multicolumn{1}{c}{\bf $SR(60)$}&\multicolumn{1}{c}{\bf $SR(80)$}
\\ \hline \\
{\bf AsymSAT} &$55.40\%$ &$27.20\%$ &$12.00\%$ &$5.00\%$ 

\\


{\bf NeuroSAT} &$33.90\%$ &$19.60\%$ &$8.50\%$ &$4.50\%$ 
\\
\end{tabular}
}
\end{center}
\end{table}

Although in this paper we mainly investigate the importance of addressing variable dependency in the end-to-end ML SAT solvers (the first category), we argue that our technique is general and  may benefit neural SAT solvers in the second category as well. For example, in NLocal-SAT, if we can provide a more accurate initial guess with the help of a tie-breaker proposed in this paper, the later stochastic local search process may be able to reach a satisfying solution with less searching effort.


\subsection{Symmetric Breaking in GNN-based SAT Solving}
Preferential Labeling \cite{sun2022generalized} is another method that can potentially break the tie between two symmetric variables in GNN-based SAT solving. It assigns distinct initial embeddings to variables, so symmetric nodes can therefore be distinguished. 
However, biased initialization also introduces artefact for GNN.
In order to smooth out the artefact, each round of training or inference must evaluate the network under multiple random permutations of the initial embeddings. In the training phase, Preferential Labeling picks the permutation that produces the lowest loss and only optimizes the network parameters under this permutation. Meanwhile, the inference process takes the averaged output among all attempted permutations as the final prediction.
Compared to Preferential Labeling,
we regard our method as a lower-cost solution for tie-breaking in SAT solving.
Our appendix details the comparison on performance and cost.\looseness=-1 









\section{Conclusion}
\label{conclusion}

This paper addresses the need of considering variable dependency when designing a machine-learning model for SAT solving. Specifically, the satisfying assignment to one variable is closely related to those made to other variables within the same SAT problem.
This paper proposes using RNNs to make sequential predictions for SAT solving. 
Our experiments show that this improvement extends the solving capability on symmetric Circuit-SAT problems and achieves a higher solution rate on randomly generated SAT and Circuit-SAT instances compared to concurrent GNN-based SAT solving methods. Although this paper focuses on the end-to-end machine-learning-based SAT solvers, using RNNs 
to account for variable dependency
may also benefit other hybrid SAT solvers that use machine learning as a guiding heuristic.\looseness=-1

\bibliographystyle{named}
\bibliography{ijcai23}
\end{document}